\documentclass{article}
\usepackage{ijcai21}

\usepackage{times}
\usepackage{soul}
\usepackage{url}
\usepackage[hidelinks]{hyperref}
\usepackage[utf8]{inputenc}
\usepackage[small]{caption}
\usepackage{graphicx}
\usepackage{amsmath}
\usepackage{amsthm}
\usepackage{booktabs}
\usepackage{algorithm}
\usepackage{algorithmic}
\usepackage{epsfig}
\usepackage{graphicx}
\usepackage{amsmath}
\usepackage{amssymb}
\usepackage{booktabs}
\usepackage{comment}
\usepackage{multirow}

\usepackage{booktabs}
\usepackage{array, caption, threeparttable}
\usepackage{caption}
\usepackage{algorithm}
\usepackage{listings}
\usepackage{color}

\usepackage{mathtools}

\usepackage{todonotes}
\usepackage{gensymb}

\title{Dual-Cross Central Difference Network for Face Anti-Spoofing}

\author{
Zitong Yu$^1$
\and
Yunxiao Qin$^2$\and
Hengshuang Zhao$^{3}$\and
Xiaobai Li$^1$\And
Guoying Zhao$^1$\thanks{Corresponding author. This work was supported by the Academy of Finland for project MiGA (grant 316765), ICT 2023 project (grant 328115), Infotech Oulu, project 6+E (grant 323287) funded by Academy of Finland, and project PhInGAIN (grant 200414) funded by The Finnish Work Environmental Fund. The authors wish to acknowledge CSC-IT Center for Science, Finland, for computational resources.}
\affiliations
$^1$CMVS, University of Oulu \quad  
$^2$Northwestern Polytechnical University \quad 
$^3$University of Oxford
\emails
\{zitong.yu, xiaobai.li, guoying.zhao\}@oulu.fi,
qyxqyx@mail.nwpu.edu.cn,
hengshuang.zhao@eng.ox.ac.uk
}

\begin{document}

\maketitle

\begin{abstract}
  
Face anti-spoofing (FAS) plays a vital role in securing face recognition systems. Recently, central difference convolution (CDC) ~\cite{yu2020searching} has shown its excellent representation capacity for the FAS task via leveraging local gradient features. However, aggregating central difference clues from all neighbors/directions simultaneously makes the CDC redundant and sub-optimized in the training phase. In this paper, we propose two Cross Central Difference Convolutions (C-CDC), which exploit the difference of the center and surround sparse local features from the horizontal/vertical and diagonal directions, respectively. It is interesting to find that, with only five ninth parameters and less computational cost, C-CDC even outperforms the full directional CDC. Based on these two decoupled C-CDC, a powerful Dual-Cross Central Difference Network (DC-CDN) is established with Cross Feature Interaction Modules (CFIM) for mutual relation mining and local detailed representation enhancement. Furthermore, a novel Patch Exchange (PE) augmentation strategy for FAS is proposed via simply exchanging the face patches as well as their dense labels from random samples. Thus, the augmented samples contain richer live/spoof patterns and diverse domain distributions, which benefits the intrinsic and robust feature learning. Comprehensive experiments are performed on four benchmark datasets with three testing protocols to demonstrate our state-of-the-art performance.
\end{abstract}

\section{Introduction}

Face recognition technology has widely used in many interactive intelligent systems due to their convenience and remarkable accuracy. However, face recognition systems are still vulnerable to presentation attacks (PAs) ranging from print, replay and 3D-mask attacks. Therefore, both the academia and industry have recognized the critical role of face anti-spoofing (FAS) for securing the face recognition system.

In the past decade, both traditional~\cite{Pereira2012LBP,Komulainen2014Context,Patel2016Secure} and deep learning-based~\cite{yu2020searching,yu2020face,Liu2018Learning,jourabloo2018face,yang2019face,yu2020multi} methods have shown effectiveness for presentation attack detection (PAD). On one hand, some classical local descriptors (e.g., local binary pattern (LBP)~\cite{boulkenafet2015face} and histogram of gradient (HOG)~\cite{Komulainen2014Context}) are robust for describing the detailed invariant information (e.g., color texture, moir$\rm\acute{e}$ pattern and noise artifacts) from spoofing faces. However, the shallow and coarse feature extraction procedure limits the discriminative capacity of these local descriptors.

On the other hand, convolutional neural networks (CNNs) focus on representing deeper semantic features to distinguish the bonafide and PAs, which are weak in capturing fine-grained intrinsic patterns (e.g., lattice artifacts) between the live and spoof faces, and easily influenced under diverse scenarios. Although central difference convolution (CDC) ~\cite{yu2020searching} introduced the central-oriented local gradient features to enhance models' generalization and discrimination capacity, it still suffers from two disadvantages. In CDC, central gradients from all neighbors are calculated, which is 1) inefficient in both inference and back-propagation stages; and 2) redundant and sub-optimized due to the discrepancy among diverse gradient directions. Thus, to study the impacts and relations among central gradients is not a trivial work.

One key challenge in the FAS task is how to learn representation with limited data as existing FAS datasets (e.g., OULU-NPU~\cite{Boulkenafet2017OULU} and SiW-M~\cite{liu2019deep}) do not have large amount of training data due to the high collection cost for both spoofing generation and video recording. Although generic augmentation manners (e.g., horizontal flip, color jitter and Cutout~\cite{devries2017improved}) are able to expand the scale and diversity of live/spoof samples, it still contributes not much performance improvement. Thus, it is worth rethinking the augmentation for FAS and design task-dedicated augmentation paradigm.

Motivated by the discussions above, we propose a novel convolution operator family called Cross Central Difference Convolution (C-CDC), which decouples the central gradient features into cross-directional combination (horizontal/vertical or diagonal) thus more efficient and concentrated for message aggregation. Furthermore, in order to
mimic more general attacks (e.g., partial print and mask attacks) and learn to distinguish spoofing in both global and patch level, Patch Exchange (PE) augmentation is proposed for mixed sample as well as corresponding dense label generation. To sum up, our contributions include:

%As shown in Fig.~\ref{fig:Figure1}, CDC is more likely to extract intrinsic spoofing patterns (e.g., lattice artifacts) than vanilla convolution in diverse environments.

 %Note that our work mainly focuses on detecting the planar attacks (e.g., print and replay).

\begin{itemize}
    \item We design a sparse convolution family called Cross Central Difference Convolution (C-CDC), which decouples the vanilla CDC into two cross (i.e., horizontal/vertical and diagonal) directions, respectively. Compared with CDC, our proposed C-CDC could achieve better performance for FAS with only five ninth parameters and less computational cost.
    
    \item  We propose a Dual-Cross Central Difference Network (DC-CDN), consisting of two-stream backbones with horizontal/vertical and diagonal C-CDC, respectively. Moreover, we also introduce Cross Feature Interaction Modules (CFIM) between two streams of DC-CDN for mutual neighbor relation mining and local detailed representation enhancement.
    
    \item We propose the first FAS-dedicated data augmentation method, Patch Exchanges (PE), to synthesize mixed samples with diverse attacks and domains, which is able to plug and play in not only DC-CDN but also existing FAS methods for performance improvement.

    \item Our proposed method achieves state-of-the-art performance on four benchmark datasets with intra-dataset, cross-dataset, and cross-type testing protocols.

\end{itemize}

\section{Related Work}

%In this section, we first introduce some recent progress in the FAS community; and then demonstrate related convolution operators in deep learning area. Finally, previous NAS methods will be reviewed.

\noindent\textbf{Face Anti-Spoofing.}\quad      
Traditional face anti-spoofing methods usually extract handcrafted features from the facial images to capture the spoofing patterns. Some classical local descriptors such as LBP~\cite{boulkenafet2015face} and
 HOG~\cite{Komulainen2014Context} are utilized for handcrafted features. More recently, a few deep learning based methods are proposed for face anti-spoofing. On the one hand, FAS can be naturally treated as a binary classification task, thus binary cross entropy loss is used for model supervision. On the other hand, according to the physical discrepancy between live and spoof faces, dense pixel-wise supervisions~\cite{yu2021revisiting} such as pseudo depth map~\cite{Liu2018Learning,yu2020searching,wang2020deep}, reflection map~\cite{yu2020face}, texture map~\cite{zhang2020face} and binary map~\cite{george2019deep} are designed for fine-grained learning. In this work, we supervise the deep networks with pseudo depth map due to its effectiveness.
 
 Due to the high collection cost for spoof attacks, there are limited data scale and diversity in public datasets. Supervised with small-scale predefined scenarios and PAs, most existing FAS methods are easy to overfit and vulnerable to domain shift and unseen attacks. In order to detect unseen attacks successfully, deep tree network~\cite{liu2019deep} and adaptive inner-update meta learning~\cite{qin2019learning} are developed for zero-shot FAS. However, it is still urgent to provide larger-scale and richer live/spoof data for deep models training. Here we consider novel data augmentation for FAS to tackle this challenge.    

%, which is not suitable to deploy in rapidly responding conditions

%\vspace{0.8em}
\noindent\textbf{Convolution Operators.}\quad  
The convolution operator is commonly used
for local feature representation in modern deep learning framework. Recently, a few extensions to the vanilla convolution operator have been proposed. In one direction, pre-defined or learnable local relation is embedded in the convolution operator. Representative works include Local Binary Convolution \cite{juefei2017local} and Gabor Convolution \cite{luan2018gabor}, which is proposed for local invariance preservation and enhancing the resistance to spatial changes, respectively. Besides, self-attention layer \cite{parmar2019stand}, self-attention block \cite{zhao2020exploring} and local relation layer \cite{hu2019local} are designed for mining the local relationship flexibly. Another direction is to modify the spatial scope for aggregation. Two related works are dialated convolution \cite{yu2015multi} and deformable convolution \cite{dai2017deformable}. However, these convolution operators may not be suitable for FAS task because of the limited representation capacity for invariant fine-grained features. In contrast, CDC~\cite{yu2020searching,yu2020fas,yu2020multi} is proposed for invariant and detailed features extraction, which is suitable for the FAS task. In this paper, we devote to improving the vanilla CDC in terms of both performance and efficiency.

\section{Methodology} \label{sec:method}
In this section we first briefly review the CDC~\cite{yu2020searching}, and then introduce the novel Cross Central Difference Convolution (C-CDC) family in Sec.~\ref{sec:CCDC}. Based on the C-CDC operators, we propose Dual-Cross Central Difference Networks in Sec.~\ref{sec:DCCDN}. Finally we present the novel data augmentation strategy in Sec.~\ref{sec:PE}.

\subsection{Preliminary}

As the basic operator in deep networks, the vanilla 2D convolution consists of two main steps: 1) \textsl{sampling} local neighbor region $\mathcal{R}$ over the input feature map $x$; and then 2) \textsl{aggregating} the sampled values via learnable weights $w$. As a result, the output feature map $y$ can be formulated as

%\vspace{-0.5em}
\begin{equation} 
y(p_0)=\sum_{p_n\in \mathcal{R}}w(p_n)\cdot x(p_0+p_n),
\label{eq:vanilla}
%\vspace{-0.5em}
\end{equation}
where $p_0$ denotes the current location on both input and output feature maps while $p_n$ enumerates the locations in $\mathcal{R}$. For instance, local receptive field region for convolution operator with 3$\times$3 kernel and dilation 1 is $\mathcal{R}=\left \{  (-1,-1),(-1,0),\cdots,(0,1),(1,1)  \right \}$.

Different from the vanilla convolution, the CDC introduces central gradient features to enhance the representation and generalization capacity, which can be formulated as
%defined in Eq.~(\ref{eq:central}).  %Eq.~(\ref{eq:vanilla}) becomes    
\begin{equation} 
%y(p_0)=\sum_{p_n\in \mathcal{R}, p_n\neq (0,0)}w(p_n)\cdot (x(p_0+p_n)-x(p_0)),
y(p_0)=\sum_{p_n\in \mathcal{R}}w(p_n)\cdot (x(p_0+p_n)-x(p_0)).
\label{eq:CCDC}
%\vspace{-0.2em}
\end{equation}
%When $p_n= (0,0)$, the gradient value always equals to zero with respect to the central location $p_0$ itself.

As both intensity-level semantic information and gradient-level detailed clues are crucial for robust FAS, the generalized CDC operator can be represented by combination of vanilla convoluiton and CDC

\begin{equation} 
\begin{split}
y(p_0)
&=\theta \cdot \underbrace{\sum_{p_n\in \mathcal{R}}w(p_n)\cdot (x(p_0+p_n)-x(p_0))}_{\text{central difference convolution}}\\
&\quad+ (1-\theta)\cdot \underbrace{\sum_{p_n\in \mathcal{R}}w(p_n)\cdot x(p_0+p_n)}_{\text{vanilla convolution}}, \\
&= \underbrace{\sum_{p_n\in \mathcal{R}}w(p_n)\cdot x(p_0+p_n)}_{\text{vanilla convolution}}+\theta\cdot (\underbrace{-x(p_0)\cdot\sum_{p_n\in \mathcal{R}}w(p_n))}_{\text{central difference term}},\\
\end{split}
\label{eq:CCDC2}
\end{equation}
where hyperparameter $\theta \in [0,1]$ trade-offs the contribution between intensity-level and gradient-level information. The higher value of $\theta$ means the more importance of central gradient features. Please note that $w(p_n)$ is shared between vanilla convolution and CDC, thus no extra parameters are added. Henceforth, the generalized CDC will be referred as \textbf{CDC} directly for clear statement.

\begin{figure}
\centering
\includegraphics[scale=0.33]{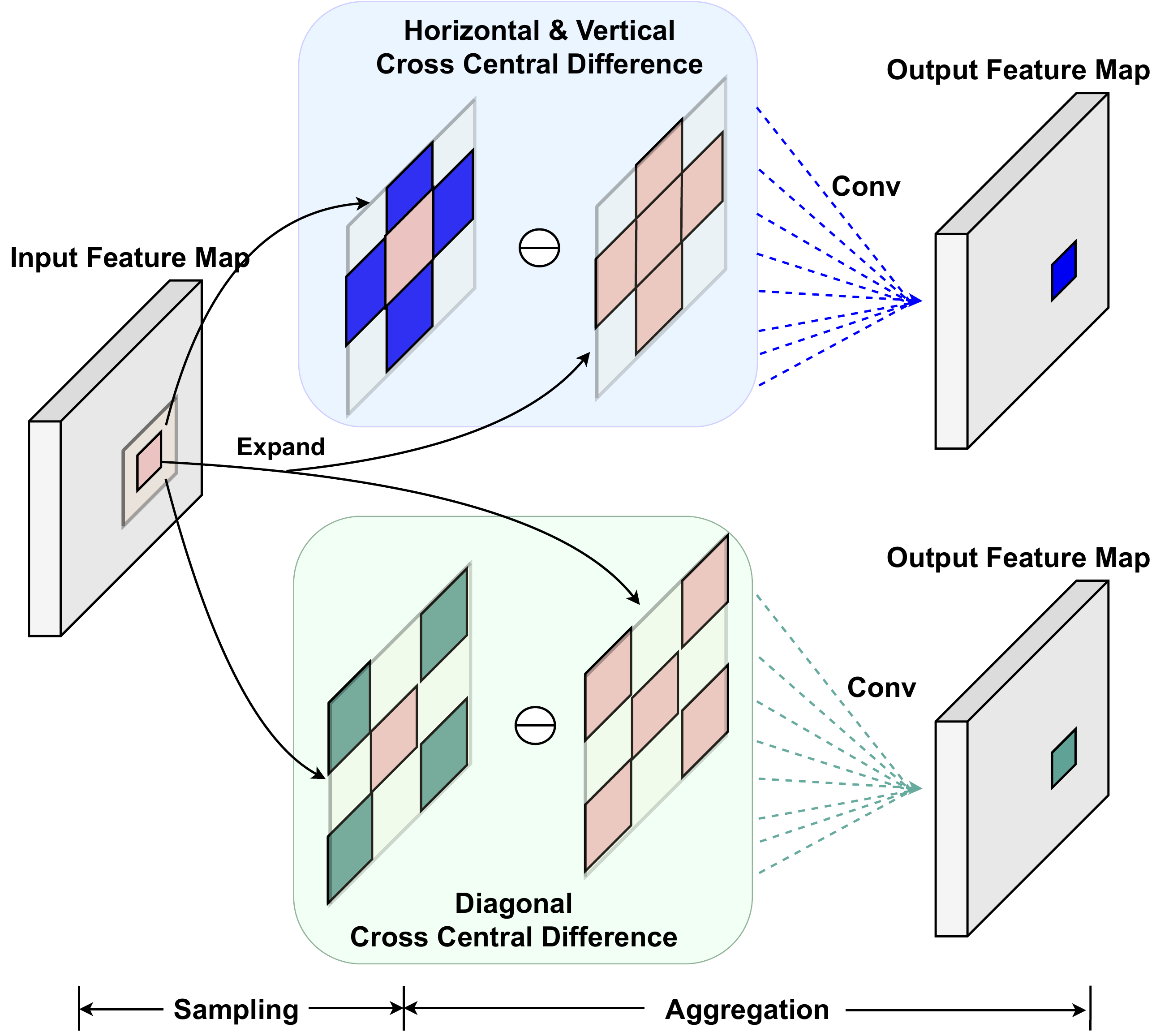}
  \caption{\small{
  Cross central difference convolution. The C-CDC(HV) in the upper part calculates the central gradients from the horizontal \& vertical neighbors while the C-CDC(DG) in the lower part from the diagonal neighbors.
  }
  }
 
\label{fig:CCDC}
\vspace{-0.5em}
\end{figure}

\subsection{Cross Central Difference Convolution}
\label{sec:CCDC}

As can be seen from Eq.~(\ref{eq:CCDC2}) that CDC aggregates both the vanilla and central gradient features from entire local neighbor region $\mathcal{R}$, which might be redundant and hard to be optimized. We assume that exploiting sparse center-oriented difference features would alleviate local competition and overfitting issues. Therefore, we propose the sparse Cross Central Difference Convolution (C-CDC) family, intending to learn more concentrated and intrinsic features for FAS. 

Compared with CDC operating on $\mathcal{R}$, C-CDC prefers to sample a sparser local region $\mathcal{S}$, which can be formulated as
\begin{equation} 
%y(p_0)=\sum_{p_n\in \mathcal{R}, p_n\neq (0,0)}w(p_n)\cdot (x(p_0+p_n)-x(p_0)),
y(p_0)=\sum_{p_n\in \mathcal{S}}w(p_n)\cdot (x(p_0+p_n)-x(p_0)).
\label{eq:CCDC3}
%\vspace{-0.2em}
\end{equation}
To be specific, we decouple $\mathcal{R}$ into two cross neighbor regions, including 1) horizontal \& vertical (HV) cross neighbor regions $\mathcal{S}_{HV}=\left \{  (-1,0),(0,-1),(0,0),(0,1),(1,0) \right \}$; and 2) diagonal (DG) cross neighbor regions $\mathcal{S}_{DG}=\left \{  (-1,-1),(-1,1),(0,0),(1,-1),(1,1) \right \}$. In this way, horizontal \& vertical C-CDC and diagonal C-CDC can be represented when $\mathcal{S}=\mathcal{S}_{HV}$ and $\mathcal{S}=\mathcal{S}_{DG}$, respectively. Figure~\ref{fig:CCDC} illustrates the workflow of the C-CDC(HV) and C-CDC(DG). Similarly, the generalized C-CDC can be easily formulated when replacing $\mathcal{R}$ with $\mathcal{S}$ in Eq.~(\ref{eq:CCDC2}). We will use this generalized C-CDC henceforth. 

In terms of designs of the sparse local region $\mathcal{S}$, there are also some other solutions with different neighbor locations or fewer neighbors. The reason that we consider the cross (i.e., HV and DG) fashions for $\mathcal{S}$ derives from their symmetry, which is beneficial for model convergence and robust feature representation. The studies about the sparsity and neighbor locations are shown in Section~\ref{sec:Ablation} and \textsl{\textbf{Appendix A}}.

\newcommand{\tabincell}[2]{\begin{tabular}{@{}#1@{}}#2\end{tabular}}
\begin{table}\footnotesize 
    
\resizebox{0.48\textwidth}{!}{
\begin{tabular}{c|c|c|c}
    \toprule[1pt]
	%\hline
	Output & DepthNet & CDCN  & C-CDN (Ours) \\
	%($\theta=0.7$)  & CDCN ($\theta=0.7$)\\
	\hline 
	 $256\times 256$ & $3\times 3 \textrm{ conv}, 64$ & $3\times 3 \textrm{ CDC}, 64$  & $3\times 3 \textbf{ C-CDC}, 64$\\
	 
	 \hline 
	    \tabincell{c}{$128\times 128$\\ (Low)}  & $\begin{bmatrix*}[l]
        3\times 3 \textrm{ conv}, 128\\ 
        3\times 3 \textrm{ conv}, 196\\ 
        3\times 3 \textrm{ conv}, 128\\ 
        3\times 3 \textrm{ max pool}
        \end{bmatrix*}$
        & $\begin{bmatrix*}[l]
        3\times 3\textrm{ CDC}, 128\\ 
        3\times 3\textrm{ CDC}, 196\\ 
        3\times 3\textrm{ CDC}, 128\\ 
        3\times 3\textrm{ max pool}
        \end{bmatrix*}$   
        & $\begin{bmatrix*}[l]
        3\times 3\textbf{ C-CDC}, 128\\ 
        3\times 3\textbf{ C-CDC}, 196\\ 
        3\times 3\textbf{ C-CDC}, 128\\ 
        3\times 3\textrm{ max pool}
        \end{bmatrix*}$\\
	 
	 \hline 
	    \tabincell{c}{$64\times 64$ \\ (Mid)}
	    &$\begin{bmatrix*}[l]
        3\times 3 \textrm{ conv}, 128\\ 
        3\times 3 \textrm{ conv}, 196\\ 
        3\times 3 \textrm{ conv}, 128\\ 
        3\times 3 \textrm{ max pool}
        \end{bmatrix*}$ 
        & $\begin{bmatrix*}[l]
        3\times 3 \textrm{ CDC}, 128\\ 
        3\times 3 \textrm{ CDC}, 196\\ 
        3\times 3 \textrm{ CDC}, 128\\ 
        3\times 3 \textrm{ max pool}
        \end{bmatrix*}$  
        & $\begin{bmatrix*}[l]
        3\times 3 \textbf{ C-CDC}, 128\\ 
        3\times 3 \textbf{ C-CDC}, 196\\ 
        3\times 3 \textbf{ C-CDC}, 128\\ 
        3\times 3 \textrm{ max pool}
        \end{bmatrix*}$ \\
        
	 \hline 
	    \tabincell{c}{$32\times 32$ \\ (High)}
	    & $\begin{bmatrix*}[l]
        3\times 3 \textrm{ conv}, 128\\ 
        3\times 3 \textrm{ conv}, 196\\ 
        3\times 3 \textrm{ conv}, 128\\ 
        3\times 3 \textrm{ max pool}
        \end{bmatrix*}$ 
        & $\begin{bmatrix*}[l]
        3\times 3 \textrm{ CDC}, 128\\ 
        3\times 3 \textrm{ CDC}, 196\\ 
        3\times 3 \textrm{ CDC}, 128\\ 
        3\times 3 \textrm{ max pool}
        \end{bmatrix*}$   
        & $\begin{bmatrix*}[l]
        3\times 3 \textbf{ C-CDC}, 128\\ 
        3\times 3 \textbf{ C-CDC}, 196\\ 
        3\times 3 \textbf{ C-CDC}, 128\\ 
        3\times 3 \textrm{ max pool}
        \end{bmatrix*}$ \\
    
    \hline 
     $32\times 32$ & \multicolumn{3}{c}{[concat (Low, Mid, High), $384$]}  \\
    
    \hline 
	    $32\times 32$ & $\begin{bmatrix*}[l]
        3\times 3 \textrm{ conv}, 128\\ 
        3\times 3 \textrm{ conv}, 64\\ 
        3\times 3 \textrm{ conv}, 1 
        \end{bmatrix*}$ 
        & $\begin{bmatrix*}[l]
        3\times 3 \textrm{ CDC}, 128\\ 
        3\times 3 \textrm{ CDC}, 64\\ 
        3\times 3 \textrm{ CDC}, 1
        \end{bmatrix*}$   
         & $\begin{bmatrix*}[l]
        3\times 3 \textbf{ C-CDC}, 128\\ 
        3\times 3 \textbf{ C-CDC}, 64\\ 
        3\times 3 \textbf{ C-CDC}, 1
        \end{bmatrix*}$ \\
    
    \hline 
    \# params &  $2.25\times 10^6$ &  $2.25\times 10^6$ &  $1.25\times 10^6$\\
    
     \hline 
    \# FLOPs &  $4.8\times10^{10}$ &  $4.8\times10^{10}$ &  $1.78\times10^{8}$ \\
	\bottomrule[1pt]
\end{tabular}
}
\caption{Architectures of DepthNet, CDCN, and the proposed C-CDN. Inside the brackets are the filter sizes and feature dimensionalities. `conv' suggests the vanilla convolution. All convolutional layers are with stride=1 and are followed by a BN-ReLU layer while pooling layers are with stride=2.}	
%\vspace{-0.9em}
\label{tab:network}
\end{table}

\subsection{Dual-Cross Central Difference Network }
\label{sec:DCCDN}

Pseudo depth-based supervision takes advantage of the discrimination between live and spoof faces based on 3D shape, which is able to provide pixel-wise detailed clues to enforce FAS model to capture intrinsic features. Following the similar depth-supervised backbone as ``DepthNet"~\cite{Liu2018Learning} and ``CDCN"~\cite{yu2020searching}, we replace all the $3\times3$ convolution operators with our proposed C-CDC to form the Cross Central Difference Network (C-CDN). Given a single RGB face image with size $3 \times 256 \times 256$, multi-level (low-level, mid-level and high-level) fused features are extracted for predicting the grayscale facial depth with size $32 \times 32$. The details of C-CDN are shown in Table~\ref{tab:network}. It can be seen that with the similar architecture (e.g., network depth and width), C-CDN has only five ninth parameters and two hundredth computational cost compared with DepthNet and CDCN due to the sparse local sampling mechanism in C-CDC. We use $\theta=0.8$ as the default setting, and the corresponding study about $\theta$ will be discussed in Section~\ref{sec:Ablation}.

\begin{figure}
\centering
\includegraphics[scale=0.24]{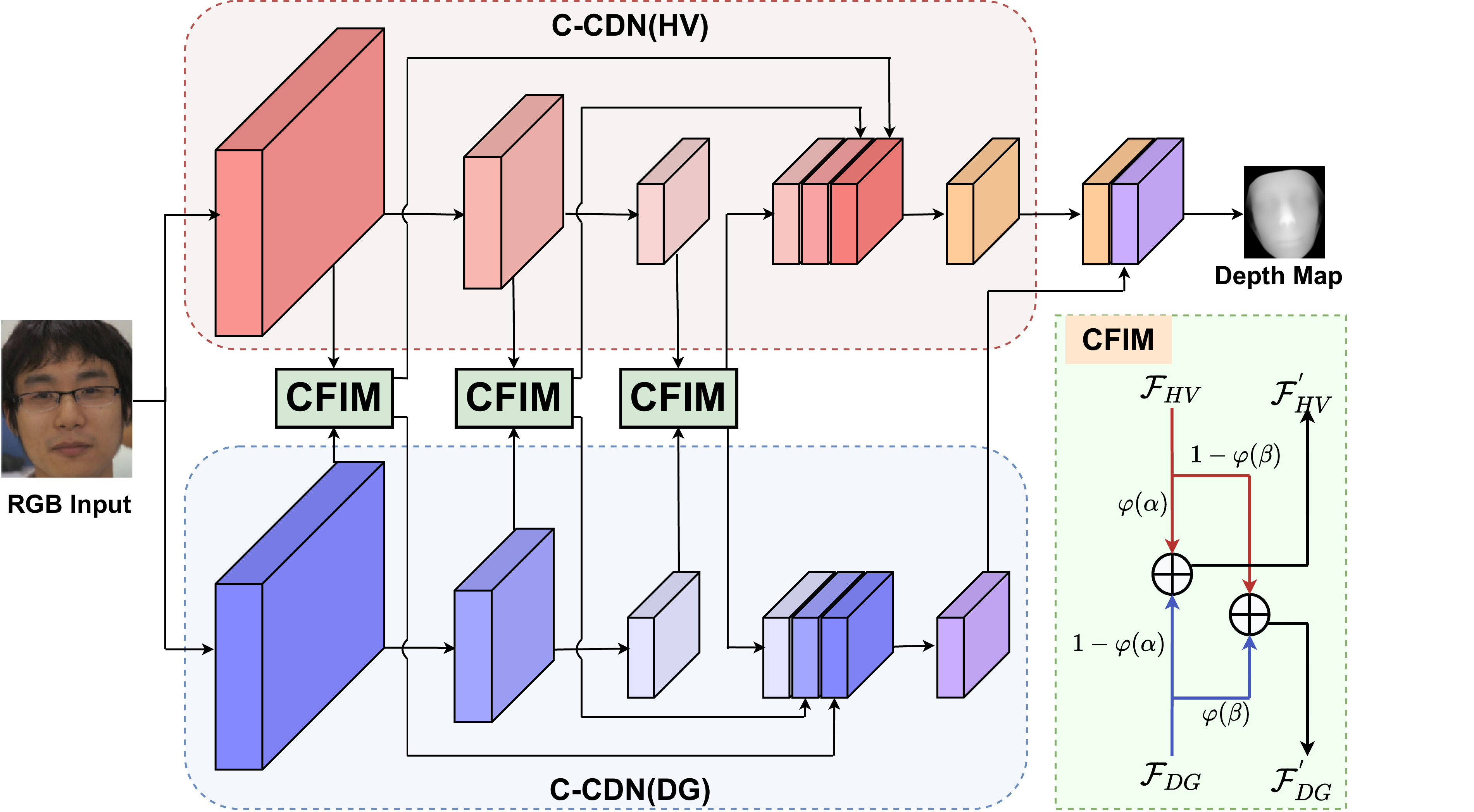}
  \caption{\small{
  Dual-cross central difference network with Cross Feature Interaction Modules for low-mid-high level feature enhancement.  
  }
  }
 
\label{fig:DCCDN}
\vspace{-0.5em}
\end{figure}

Although the C-CDC decouples and learns the local gradient features with particular views (HV and DG), it still suffers from information loss compared with CDC operating on the full local neighbors. In order to fully exploit the local features and interact between HV and DG views, a Dual-Cross Central Difference Network (DC-CDN)\footnote{\href{https://github.com/ZitongYu/CDCN}{https://github.com/ZitongYu/CDCN}} is proposed. As shown in Figure~\ref{fig:DCCDN}, two independent (unshared) networks respectively assembled with C-CDC(HV) and C-CDC(DG) are used. Then the extracted dual-stream features from different views are fused for final depth prediction. In this way, the full neighbor aggregation step can be disentangled into two sub-steps: 1) sparse neighbor aggregation for individual stream; and 2) dual-stream fusion.

\noindent\textbf{Cross Feature Interaction Module.}\quad With only simple late fusion, the performance improvement might be limited due to the lack of message passing from the preceding (i.e., low-level, mid-level, and high-level) stages. In order to effectively mine the relations across dual streams and enhance local detailed representation capacity, we propose the Cross Feature Interaction Module (CFIM) to fuse dual-stream multi-level features adaptively. To be specific, given the two-stream features $\mathcal{F}_{HV}$ and $\mathcal{F}_{DG}$, the CFIM enhanced features $\mathcal{F}^{'}_{HV}$ and $\mathcal{F}^{'}_{DG}$ can be formulated as 
\begin{equation} 
\begin{split}
\mathcal{F}^{'}_{HV}=\varphi (\alpha)\cdot  \mathcal{F}_{HV}+(1-\varphi (\alpha))\cdot \mathcal{F}_{DG},\\
\mathcal{F}^{'}_{DG}=\varphi (\beta)\cdot \mathcal{F}_{DG}+(1-\varphi (\beta)) \cdot\mathcal{F}_{HV},\\
\end{split}
\label{eq:CFIM}
\end{equation}
where $\varphi()$ denotes the Sigmoid function for [0,1] range mapping. $\alpha$ and $\beta$ are the attention weights for $\mathcal{F}_{HV}$ and $\mathcal{F}_{DG}$, respectively. In our default setting, both $\alpha$ and $\beta$ are initialized to 0 and learnable, which could be adaptively adjusted during the training iterations. We also investigate the manually fixed $\alpha$ and $\beta$ in \textsl{\textbf{Appendix B}}. As illustrated in Fig.~\ref{fig:DCCDN}, here we respectively plug three CFIMs (with learnable $\alpha_{low}$, $\alpha_{mid}$, $\alpha_{high}$, $\beta_{low}$, $\beta_{mid}$, $\beta_{high}$) in the output features from low-mid-high levels before multi-level concatenation.

\begin{algorithm}[t]
	\caption{Patch Exchange Augmentation}
	{\bfseries Input:} Face images $I$ with batchsize $N$, pseudo depth map labels $D$, augmented ratio $\gamma\in [0,1]$, step number $\rho$ 
	\\
	{\bfseries 1\,\,\,:} {\bfseries for } each $I_{i}$ and $D_{i}$, $i=1,...,\left \lfloor \gamma*N\right \rfloor$ {\bfseries do}\\
	% and meta-learner $\widetilde{\varphi}$ \\
	{\bfseries 2\,\,\,:}  \ \ \,   {\bfseries for} each step $\rho$ {\bfseries do} \\
	{\bfseries 3\,\,\,:}  \ \ \quad \, Randomly select a patch region $P$ within $I_{i}$ \\
	{\bfseries 4\,\,\,:}  \ \ \quad \, Randomly select a batch index $j, j\leq N$ \\
	{\bfseries 5\,\,\,:}  \ \ \quad \, Exchange the image patch $I_{i}(P)=I_{j}(P)$ and label patch $D_{i}(P)=D_{j}(P)$\\
	{\bfseries 6\,\,\,:} 	\ \ \, {\bfseries end} \\
	{\bfseries 7:} 	{\bfseries end} \\
	{\bfseries 8:} 	{\bfseries return} augmented $I$ and $D$ 
	\label{algorithm:PE}
\end{algorithm}

\begin{figure}
\centering
\includegraphics[scale=0.19]{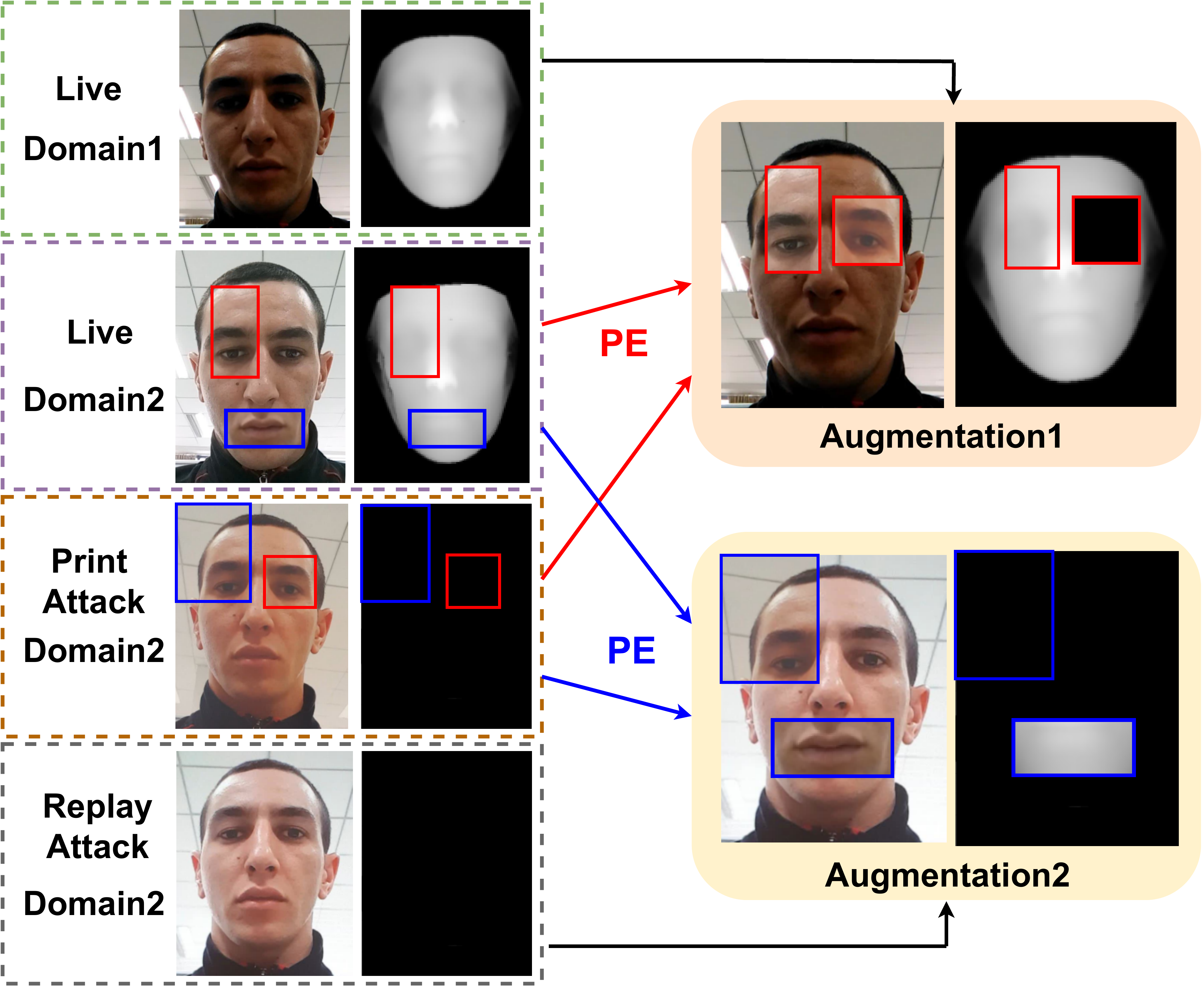}
\vspace{-0.6em}
  \caption{\small{
  Patch Exchange augmentation. Two augmented samples are synthesized via exchanging the RGB patches as well as corresponding pseudo depth labels from ``Live Domain2" and ``Print Attack Domain2". Thus, the samples ``Augmentation1" and ``Augmentation2" contain diverse domains and attack types, respectively. 
  }
  }
 
\label{fig:PE}
\vspace{-0.8em}
\end{figure}

\subsection{Patch Exchange Augmentation}
\label{sec:PE}
Due to the high collection cost for spoof attacks, there are limited data scale and diversity in public FAS datasets. In this paper, we also propose a FAS-dedicated data augmentation method, named Patch Exchanges (PE), to synthesize mixed samples with diverse attacks and domains. There are three advantages for PE augmentation: 1) face patches from different domains (e.g., recorded scenario, sensor, and subject) are introduced for enriching data distribution; 2) random live and PA patches are exchanged to mimic arbitrary partial attacks; and 3) the exchanged patches with corresponding dense labels enforce the model to learn more detailed and intrinsic features for spoofing detection. The complete algorithm of PE is summarized in Algorithm~\ref{algorithm:PE}. The two hyperparameters $\gamma$ and $\rho$ control the augmentation ratio and intensity, respectively. As a tradeoff, we use empirical settings $\gamma=0.5$ and $\rho=2$ for experiments. 

%besides the novel operators and architecture design, 

Note that as the face images $I$ for PE are coarsely aligned, the exchanged patches would have similar semantic content (e.g., cheek, nose and mouth) but with diverse live/spoof clues. Thus, the augmented live/spoof faces are still realistic and even more challenging to be distinguished. Some typical samples with PE augmentation are visualized in Figure~\ref{fig:PE}.

\vspace{-0.4em}
\section{Experiments}
\vspace{-0.1em}
\label{sec:experiemnts}
%In this section, extensive experiments are performed to demonstrate the effectiveness of our method. In the following, we sequentially describe the employed datasets \& metrics (Sec. \ref{sec:dataset}), implementation details (Sec. \ref{sec:Details}), results (Sec. \ref{sec:Ablation} - \ref{sec:Inter}) and analysis (Sec. \ref{sec:Analysis}).

\subsection{Datasets and Metrics}

\label{sec:dataset}
\textbf{Databases.} 
Four databases OULU-NPU~\cite{Boulkenafet2017OULU}, CASIA-MFSD~\cite{Zhang2012A}, Replay-Attack~\cite{ReplayAttack} and SiW-M~\cite{liu2019deep} are used in our experiments. OULU-NPU is a high-resolution database, containing four protocols to evaluate the generalization (e.g., unseen illumination and attack medium) of models respectively, which is used for intra testing. CASIA-MFSD and Replay-Attack are small-scale databases with low-resolution videos, which are used for cross testing. SiW-M is designed for cross-type testing for unseen attacks as there are rich (13) attack types inside.  

\noindent\textbf{Performance Metrics.}\quad
In OULU-NPU dataset, we follow the original protocols and metrics, i.e., Attack Presentation Classification Error Rate (APCER), Bona Fide Presentation Classification Error Rate (BPCER), and ACER~\cite{ACER} for a fair comparison. Half Total Error Rate (HTER) is adopted in the cross testing between CASIA-MFSD and Replay-Attack. For the cross-type test on SiW-M, ACER and Equal Error Rate (EER) are employed for evaluation.

\begin{figure}
\centering
\includegraphics[scale=0.35]{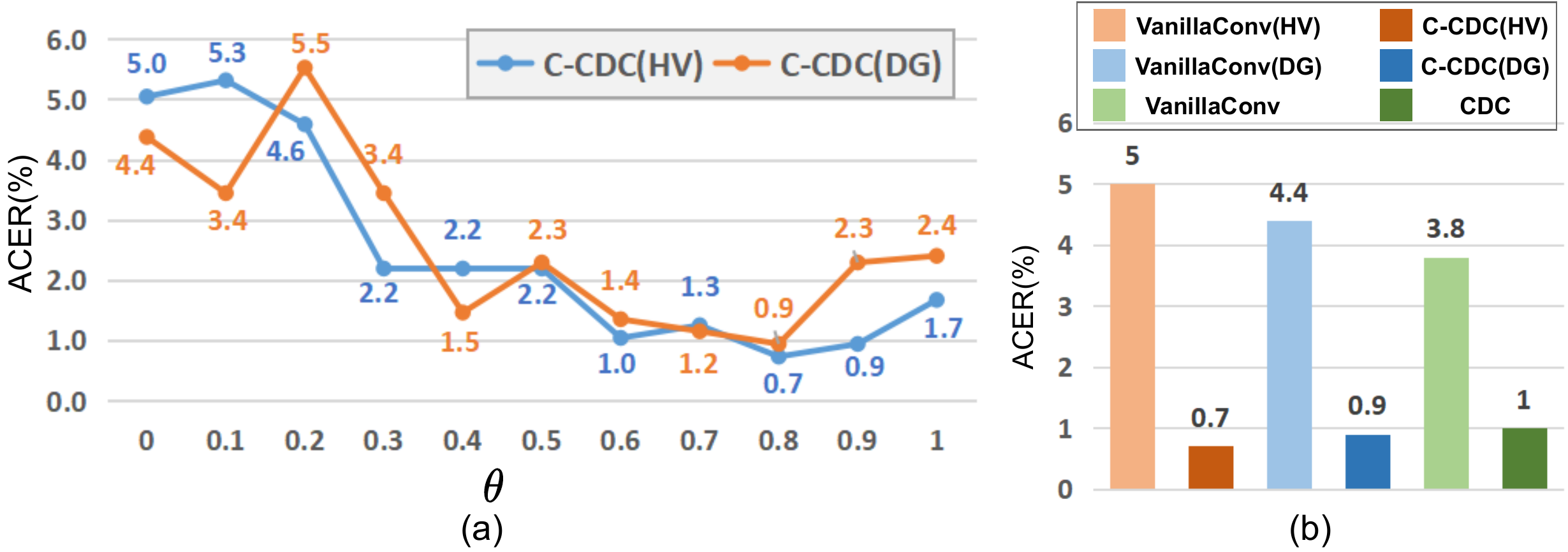}
\vspace{-2.0em}
  \caption{\small{
  (a) Impact of $\theta$ in C-CDN. (b) Comparison among various convolutions. The lower ACER, the better performance.  
  }
  }
 
\label{fig:ablation1}
\vspace{-0.5em}
\end{figure}

\begin{table}[t]\small
\centering
\caption{The ablation study about DC-CDN, CFIM, and PE augmentation on Protocol 1 of OULU-NPU.}
\vspace{-0.8em}
\scalebox{0.84}{\begin{tabular}{l|c | c|c}
\toprule[1pt]
Model & CFIM & PE Augmentation & ACER(\%)\\
\hline
C-CDN(HV) &  &  & 0.7 \\
C-CDN(DG) &  & & 0.9 \\
DC-CDN (average) &  &  & 1.0 \\
DC-CDN (concat) &  &  & 0.8 \\
DC-CDN (concat) & $\surd$ &  & 0.5 \\
\hline
C-CDN(HV) &  & $\surd$ & 0.6 \\
C-CDN(DG) &  & $\surd$  & 0.7 \\
DC-CDN (concat) & $\surd$ & $\surd$  & 0.4 \\

\bottomrule[1pt]

\end{tabular}}
\label{tab:ablation}
\vspace{-1.0em}
\end{table}

\subsection{Implementation Details}
\vspace{-0.2em}
\label{sec:Details}

\textbf{Depth Generation.}\quad
 Dense face alignment~\cite{guo2020towards} is adopted for pseudo depth generation. To clearly distinguish live and spoof faces, at the training stage, we follow~\cite{Liu2018Learning} to normalize the live depth maps in a range of $[0, 1]$, while leaving spoof depth maps to all zeros.

\noindent\textbf{Training and Testing Setting.}\quad 
Our proposed method is implemented with Pytorch. In the training stage, models are trained with batch size 8 and Adam optimizer on a single V100 GPU. Data augmentations including horizontal flip, color jitter and Cutout are used as baseline. The initial learning rate (lr) and weight decay are 1e-4 and 5e-5, respectively. We train models with maximum 800 epochs while lr halves in the 500th epoch. Similar to~\cite{yu2020searching}, all the models are supervised by mean square error (MSE) and contrastive depth loss (CDL). In the testing stage, we calculate the mean value of the predicted depth map as the final score.

\subsection{Ablation Study}
\label{sec:Ablation}
In this subsection, all ablation studies are conducted on the Protocol-1 (different illumination condition and scenario between training and testing sets) of OULU-NPU dataset.

\begin{table}[t]
\centering
\caption{The results of intra testing on the OULU-NPU dataset.}
\vspace{-0.8em}
\resizebox{0.48\textwidth}{!}{
\begin{tabular}{c|c|c|c|c}
\toprule[1pt]

Prot. & Method & APCER(\%)$\downarrow$ & BPCER(\%)$\downarrow$ & ACER(\%)$\downarrow$ \\
\hline
\multirow{6}{*}{1}
        %&GRADIANT ~\cite{boulkenafet2017competition}&1.3 &12.5 & 6.9 \\
        %&MILHP &8.3 &0.8 & 4.6 \\
        &STASN~\cite{yang2019face} &1.2 &2.5 & 1.9 \\
        &Auxiliary~\cite{Liu2018Learning} &1.6 &1.6 & 1.6 \\
        &FaceDs~\cite{jourabloo2018face} &1.2 &1.7 & 1.5 \\
        &SpoofTrace~\cite{liu2020disentangling} &0.8 &1.3 & 1.1 \\
        &Disentangled~\cite{zhang2020face} &1.7 &0.8 & 1.3 \\
        &FAS-SGTD~\cite{wang2020deep} &2.0 &0.0 & 1.0 \\
        &CDCN~\cite{yu2020searching} &0.4 &1.7 & 1.0 \\
        &BCN~\cite{yu2020face} &0.0 &1.6 & 0.8 \\
        &DeepPix~\cite{george2019deep}&0.8 &0.0 & \textbf{0.4} \\
        &\textbf{DC-CDN (Ours)}& 0.5
       &0.3 &\textbf{0.4} \\

\hline
\multirow{6}{*}{2} 
       &DeepPix~\cite{george2019deep}&11.4 &0.6 & 6.0 \\
       %&MILHP &5.6 &5.3 & 5.4 \\
       &FaceDs~\cite{jourabloo2018face}&4.2 &4.4 & 4.3 \\
       &Auxiliary~\cite{Liu2018Learning}&2.7 &2.7 & 2.7 \\
       &Disentangled~\cite{zhang2020face} &1.1 &3.6 & 2.4 \\
       &STASN~\cite{yang2019face}&4.2 &0.3 & 2.2 \\
       &BCN~\cite{yu2020face} &2.6 &0.8 & 1.7 \\
       &SpoofTrace~\cite{liu2020disentangling} &2.3 &1.6 & 1.9 \\
       &FAS-SGTD~\cite{wang2020deep} &2.5 & 1.3 & 1.9 \\
       &CDCN~\cite{yu2020searching} &1.5 &1.4 & 1.5 \\
       &\textbf{DC-CDN (Ours)} &0.7 &1.9 &\textbf{1.3} \\
\hline
\multirow{4}{*}{3} 
       &DeepPix~\cite{george2019deep}&11.7$\pm$19.6 &10.6$\pm$14.1 & 11.1$\pm$9.4 \\
       &FaceDs~\cite{jourabloo2018face}&4.0$\pm$1.8 &3.8$\pm$1.2 &3.6$\pm$1.6 \\
       &Auxiliary~\cite{Liu2018Learning}&2.7$\pm$1.3 &3.1$\pm$1.7 &{2.9}$\pm$1.5 \\
       &STASN~\cite{yang2019face}&4.7$\pm$3.9 &0.9$\pm$1.2  &2.8$\pm$1.6 \\
       &SpoofTrace~\cite{liu2020disentangling}&1.6 $\pm$1.6 & 4.0$\pm$5.4  &2.8$\pm$3.3 \\ &FAS-SGTD~\cite{wang2020deep}&3.2$\pm$2.0 & 2.2$\pm$1.4 & 2.7$\pm$0.6 \\
       &BCN~\cite{yu2020face}&2.8$\pm$2.4 &2.3$\pm$2.8  &2.5$\pm$1.1 \\
      &CDCN~\cite{yu2020searching} &2.4$\pm$1.3 &2.2$\pm$2.0  &2.3$\pm$1.4 \\
      &Disentangled~\cite{zhang2020face}&2.8$\pm$2.2 &1.7$\pm$2.6  &2.2$\pm$2.2 \\
       &\textbf{DC-CDN (Ours)} &2.2$\pm$2.8 &1.6$\pm$2.1  &\textbf{1.9$\pm$1.1} \\

\hline
\multirow{4}{*}{4} 
        &DeepPix~\cite{george2019deep}&36.7$\pm$29.7 &13.3$\pm$14.1 & 25.0$\pm$12.7 \\
       &Auxiliary~\cite{Liu2018Learning}&9.3$\pm$5.6 &10.4$\pm$6.0 &9.5$\pm$6.0 \\
       &STASN~\cite{yang2019face}&6.7$\pm$10.6 &8.3$\pm$8.4  &7.5$\pm$4.7 \\
       &CDCN~\cite{yu2020searching} &4.6$\pm$4.6 &9.2$\pm$8.0  &6.9$\pm$2.9 \\
       &FaceDs~\cite{jourabloo2018face}&1.2$\pm$6.3&6.1$\pm$5.1 &5.6$\pm$5.7 \\
       &BCN~\cite{yu2020face}&2.9$\pm$4.0 &7.5$\pm$6.9  &5.2$\pm$3.7 \\
       &FAS-SGTD~\cite{wang2020deep}&6.7$\pm$7.5 &3.3$\pm$4.1 & 5.0$\pm$2.2 \\
       &Disentangled~\cite{zhang2020face}&5.4$\pm$2.9 &3.3$\pm$6.0  &4.4$\pm$3.0 \\
     &SpoofTrace~\cite{liu2020disentangling}&2.3$\pm$3.6 &5.2$\pm$5.4  &\textbf{3.8$\pm$4.2} \\
       &\textbf{\textbf{DC-CDN (Ours)}} &5.4$\pm$3.3 &2.5$\pm$4.2  &4.0$\pm$3.1 \\
\bottomrule[1pt]
\end{tabular}
}%\resizebox{\textwidth}{!}{
\label{tab:OULU}
\vspace{-0.8em}
\end{table}

\noindent\textbf{Impact of $\theta$ in C-CDC.}\quad 
As discussed in Section~\ref{sec:CCDC}, $\theta$ controls the contribution of the gradient-based features, i.e., the higher $\theta$, the more local detailed information included. As illustrated in Fig.~\ref{fig:ablation1}(a), when $\theta\geqslant0.3$, C-CDC(HV) and C-CDC(DG) always achieve better performance than their vanilla counterpart (i.e., $\theta=0$), indicating the effectivenss of local gradient features for FAS task. As the best performance (ACER=0.7\% and 0.9\% for C-CDC(HV) and C-CDC(DG), respectively) are obtained when $\theta=0.8$, we use this setting for the following experiments. 

%Besides keeping the constant $\theta$ for all layers, we also explore an adaptive CDC method to learn $\theta$ for every layer, which is shown in \textsl{\textbf{Appendix B}}. 

\begin{table*}
\centering
\caption{Results of the cross-type testing on the SiW-M dataset.}
\vspace{-0.8em}
\scalebox{0.65}{\begin{tabular}{c|c|c|c|c|c|c|c|c|c|c|c|c|c|c|c}
\toprule[1pt]
\multirow{2}{*}{Method} &\multirow{2}{*}{Metrics(\%)} &\multirow{2}{*}{Replay} &\multirow{2}{*}{Print} &\multicolumn{5}{c|}{Mask Attacks} &\multicolumn{3}{c|}{Makeup Attacks}&\multicolumn{3}{c|}{Partial Attacks} &\multirow{2}{*}{Average} \\
\cline{5-15} &  &  &  & \tabincell{c}{Half} &\tabincell{c}{Silicone} &\tabincell{c}{Trans.} &\tabincell{c}{Paper}&\tabincell{c}{Manne.}&\tabincell{c}{Obfusc.}&\tabincell{c}{Im.}&\tabincell{c}{Cos.}&\tabincell{c}{Fun.} & \tabincell{c}{Glasses} &\tabincell{c}{Partial} & \\

%\hline

%\multirow{2}{*}{SVM+LBP~\cite{Boulkenafet2017OULU}} & ACER & 20.6 & 18.4 & 31.3 & 21.4 & 45.5 & 11.6 & 13.8 & 59.3 & 23.9 & 16.7 & 35.9 & 39.2 & 11.7 & 26.9$\pm$14.5 \\
%  & EER & 20.8 & 18.6 & 36.3  & 21.4 & 37.2 & 7.5 & 14.1 & 51.2 & 19.8 & 16.1 & 34.4 & 33.0 & 7.9 & 24.5$\pm$12.9 \\

\hline

\multirow{2}{*}{Auxiliary~\cite{Liu2018Learning}} &  ACER & 16.8 & 6.9 & 19.3 & 14.9 & 52.1 & 8.0 & 12.8 & 55.8 & 13.7 & 11.7 & 49.0 & 40.5 & 5.3 & 23.6$\pm$18.5 \\
  & EER & 14.0 & 4.3 & 11.6  & 12.4 & 24.6 & 7.8 & 10.0 & 72.3 & 10.1 & \textbf{9.4} & 21.4 & 18.6 & 4.0 & 17.0$\pm$17.7 \\

\hline

\multirow{2}{*}{DTN~\cite{liu2019deep}}  & ACER & 9.8 & 6.0 & 15.0 & 18.7 & 36.0 & 4.5 & 7.7 & 48.1 & 11.4 & 14.2 & \textbf{19.3} & 19.8 & 8.5 & 16.8 $\pm$11.1 \\
 & EER & 10.0 & \textbf{2.1} & 14.4 & 18.6 & 26.5 & 5.7 & 9.6 & 50.2 & 10.1 & 13.2 & 19.8 & 20.5 & 8.8 & 16.1$\pm$ 12.2 \\

\hline 
\multirow{2}{*}{CDCN~\cite{yu2020searching}} & ACER & 8.7 & 7.7 & 11.1 & 9.1 & 20.7 & 4.5 & 5.9 & 44.2 & 2.0 & 15.1 & 25.4 & 19.6 & 3.3 & 13.6 $\pm$11.7 \\
 & EER & 8.2 & 7.8 & 8.3 & \textbf{7.4} & 20.5 & 5.9 & 5.0 & 47.8 & 1.6 & 14.0 & 24.5 & 18.3 & \textbf{1.1} & 13.1$\pm$ 12.6 \\

\hline

\multirow{2}{*}{SpoofTrace ~\cite{liu2020disentangling}}  & ACER & \textbf{7.8} & 7.3 & \textbf{7.1} & 12.9 & \textbf{13.9} & 4.3 & 6.7 & 53.2 & 4.6 & 19.5 & 20.7 & 21.0 & 5.6 & 14.2 $\pm$13.2\\
 & EER & \textbf{7.6} & 3.8 & 8.4 & 13.8 & 14.5 & \textbf{5.3} & 4.4 & 35.4 & \textbf{0.0} & 19.3 & 21.0 & 20.8 & 1.6 & 12.0$\pm$ 10.0 \\

\hline

\multirow{2}{*}{BCN ~\cite{yu2020face}}  & ACER & 12.8 & \textbf{5.7} & 10.7 & 10.3 & 14.9 & \textbf{1.9} & 2.4 & \textbf{32.3} & 0.8 & 12.9 & 22.9 & 16.5 & \textbf{1.7} & \textbf{11.2 $\pm$9.2}\\
 & EER & 13.4 & 5.2 & \textbf{8.3} & 9.7 & 13.6 & 5.8 & 2.5 & \textbf{33.8} & \textbf{0.0} & 14.0 & 23.3 & 16.6 & 1.2 & 11.3$\pm$ 9.5 \\

\hline
\hline

\multirow{2}{*}{\textbf{DC-CDN w/o PE (Ours)}} & ACER & 12.9 & 10.2 & 9.7 & 9.1 & 16.5 & 5.3 & \textbf{1.6} & 44.6 & 0.8 & 14.0 & 22.9 & 17.3 & 3.8 & 12.9 $\pm$11.6 \\
 & EER & 12.3 & 8.7 & 12.6 & \textbf{7.4} & 13.6 & 5.9 & \textbf{0.0} & 43.4 & \textbf{0.0} & 14.0 & 19.5 & 16.7 & 2.3 & 12.0$\pm$ 11.3 \\

\hline

\multirow{2}{*}{\textbf{DC-CDN (Ours)}}  & ACER & 12.1 & 9.7 & 14.1 & \textbf{7.2} & 14.8 & 4.5 & \textbf{1.6} & 40.1 & \textbf{0.4} & \textbf{11.4} & 20.1 & \textbf{16.1} & 2.9 & 11.9$\pm$10.3 \\
  & EER & 10.3 & 8.7 & 11.1  & \textbf{7.4} & \textbf{12.5} & 5.9 & \textbf{0.0} & 39.1 & \textbf{0.0} & 12.0 & \textbf{18.9} & \textbf{13.5} & 1.2 & \textbf{10.8$\pm$10.1} \\

\bottomrule[1pt]

\end{tabular}
}
\vspace{-1.0em}
\label{tab:SiW-M}
\end{table*}

\begin{table}
\centering
\caption{The results of cross-dataset testing between CASIA-MFSD and Replay-Attack. The evaluation metric is HTER(\%).}
\vspace{-0.8em}
%\resizebox{0.48\textwidth}{!}
\scalebox{0.72}{\begin{tabular}{c|c|c|c|c}
\toprule[1pt]
\multirow{2}{*}{Method} &Train &Test &Train &Test\\
\cline{2-3} \cline{4-5} &\tabincell{c}{CASIA-\\MFSD} &\tabincell{c}{Replay-\\Attack} &\tabincell{c}{Replay-\\Attack} &\tabincell{c}{CASIA-\\MFSD}\\
\hline

%ColTexture~\cite{Boulkenafet2017Face}
%&\multicolumn{2}{c|}{30.3} &\multicolumn{2}{c|}{37.7} \\
FaceDs~\cite{jourabloo2018face}
&\multicolumn{2}{c|}{28.5} &\multicolumn{2}{c}{41.1} \\

STASN~\cite{yang2019face}
&\multicolumn{2}{c|}{31.5} &\multicolumn{2}{c}{30.9} \\

Auxiliary~\cite{Liu2018Learning}
&\multicolumn{2}{c|}{27.6} &\multicolumn{2}{c}{28.4} \\
Disentangled~\cite{zhang2020face}
&\multicolumn{2}{c|}{22.4} &\multicolumn{2}{c}{30.3} \\
FAS-SGTD~\cite{wang2020deep}
&\multicolumn{2}{c|}{17.0} &\multicolumn{2}{c}{\textbf{22.8}} \\
BCN~\cite{yu2020face}
&\multicolumn{2}{c|}{16.6} &\multicolumn{2}{c}{36.4} \\
CDCN~\cite{yu2020searching}
&\multicolumn{2}{c|}{15.5} &\multicolumn{2}{c}{32.6} \\

\hline

\textbf{DC-CDN w/o PE (Ours)}
&\multicolumn{2}{c|}{8.5} &\multicolumn{2}{c}{32.3} \\
\textbf{DC-CDN (Ours)}
&\multicolumn{2}{c|}{\textbf{6.0}} &\multicolumn{2}{c}{30.1} \\
\bottomrule[1pt]
\end{tabular}
}%\resizebox{\textwidth}{!}{
\label{tab:cross-testing}
\vspace{-0.8em}
\end{table}

\noindent\textbf{C-CDC vs. CDC.}\quad 
%As discussed in Sec. \ref{sec:CDC} about the relation between CDC and prior convolutions, we argue that the proposed CDC is more suitable for FAS task as the detailed spoofing artifacts in diverse environments should be represented by the gradient-based invariant features.
Here we evaluate the impacts of the neighbor sparsity for both vanilla and gradient-based convolutions. 
As shown in Fig.~\ref{fig:ablation1}(b), ``VanillaConv(HV)" and ``VanillaConv(DG)" performs more poorly than ``VanillaConv", which might be caused by the limited semantic representation capacity from the sparse neighbor sampling. In contrast, it is interesting to see that their central difference-based counterparts perform inversely. Compared with CDC, C-CDC(HV) and C-CDC(DG) have only five ninth parameters, and even achieve better performance (-0.3\% and -0.1\% ACER, respectively), indicating the efficiency of cross gradient features for FAS. More discussion about the sparsity and neighbor locations will be shown in \textsl{\textbf{Appendix A}}.

\noindent\textbf{Effectiveness of DC-CDN and CFIM.}\quad 
The upper part of Table~\ref{tab:ablation} shows the ablation results of DC-CDN w/ and w/o CFIM. It can be seen from the third and fourth rows that DC-CDN w/o CFIM achieves worse when using simple final depth map averaging or late fusion via concatenation. In contrast, assembled with CFIM, DC-CDN obtains 0.2\% and 0.4\% ACER decrease compared with one-stream C-CDN(HV) and C-CDN(DG), respectively. It demonstrates the importance of adaptive mutual feature interaction during the training stage. More discussion about the MFIM and the learned $\alpha,\beta$ visualization are shown in \textsl{\textbf{Appendix B}}.

\noindent\textbf{Effectiveness of PE augmentation.}\quad 
It can be seen from the lower part of Table~\ref{tab:ablation} that PE augmentation improves the performance of both single-stream (C-CDN) and dual-stream (DC-CDN) models on Protocol 1 (with slight domain shifts) of OULU-NPU. It is worth noting that, benefited from the augmented data with rich and diverse domain and attack clues, DC-CDN could obtain remarkable performance gains on more challenging senarios, including cross-type (-1.2\% EER, see Table~\ref{tab:SiW-M}) and cross-dataset (respective -2.5\% and -2.2\% HTER, see Table~\ref{tab:cross-testing}) testings. We also evaluate the generalization of PE for other architectures in \textsl{\textbf{Appendix C}}.

 \vspace{-0.2em}
\subsection{Comparison with State of the Arts}

\noindent\textbf{Intra Testing on OULU-NPU.} \quad  As shown in Table~\ref{tab:OULU}, our proposed DC-CDN ranks first on the first three protocols (0.4\%, 1.3\% and 1.9\% ACER, respectively), which indicates the proposed method performs well at the generalization of the external environment, attack mediums and input camera variation. It is clear that the proposed DC-CDN consistently outperforms CDCN~\cite{yu2020searching} on all protocols with -0.6\%, -0.2\%, -0.4\%, and -2.9\%, respectively, indicating the superiority of DC-CDN. In the most challenging Protocol 4, DC-CDN also achieves comparable performance with state-of-the-art SpoofTrace~\cite{liu2020disentangling} but more robustness with smaller ACER standard deviation among six kinds of unseen scenarios.

%To further testify the generalization ability of our models, we conduct cross-type and cross-dataset testings under unknown PAs and unseen environment, respectively.

\noindent\textbf{Cross-type Testing on SiW-M.} \quad  Following the leave-one-type-out (total 13 attack types) protocol on SiW-M, we compare the proposed methods with several recent FAS methods to validate the generalization capacity of unseen attacks. As shown in Table~\ref{tab:SiW-M}, our method achieves the best EER performance and can perform more robustly in several challenging attacks including `Silicone Mask', `Transparent Mask', `Mannequin', `Impersonation', `Funny Eye' and `Paper Glasses'. Note that our DC-CDN could achieve comparable performance with the state-of-the-art method BCN~\cite{yu2020face}, which needs the supervision signals from three kinds of pixel-wise labels (i.e., binary mask, pseudo depth and reflection map). Moreover, it is reasonable to see from the last two rows of Table~\ref{tab:SiW-M} that PE augmentation helps to improve generalization ability of unknown Partial Attacks (i.e., Funny Eye, Paper Glasses and Partial Paper) obviously.

\noindent\textbf{Cross-dataset Testing.} \quad   Here cross-dataset testing is conducted to further testify the generalization ability of our models under unseen scenarios. There are two cross-dataset testing protocols. One is that training on the CASIA-MFSD and testing on Replay-Attack, which is named as protocol CR; the second one is exchanging the training dataset and the testing dataset, named protocol RC. As shown in Table~\ref{tab:cross-testing}, our proposed DC-CDN achieves 6.0\% HTER on protocol CR, outperforming the prior state-of-the-art by a convincing margin of 9.5\%. For protocol RC, we also slightly outperform state-of-the-art frame-level methods (e.g., CDCN, BCN and Disentangled~\cite{zhang2020face}). Note that . Inspired by the dynamic clues usage in Auxiliary~\cite{Liu2018Learning} and FAS-SGTD~\cite{wang2020deep}, the performance of DC-CDN might be further boosted via introducing the similar temporal dynamic features, which will be explored in our future work.

\vspace{-0.8em}
\section{Conclusions}
\vspace{-0.3em}
\label{sec:conc}

In this paper, we propose a sparse operator family called Cross Central Difference Convolution (C-CDC) for face anti-spoofing task. Moreover, we design a Dual-Cross Central Difference Network with Cross Feature Interaction Modules for dual-stream feature enhancement. Besides, a novel Patch Exchange augmentation strategy is proposed for enriching the training data. Extensive experiments are performed to verify the effectiveness of the proposed methods. In the future, we will extend the C-CDC family to a 3D version for fine-grained video understanding tasks.

\section*{Appendix}

\begin{figure}
\centering
\includegraphics[scale=0.32]{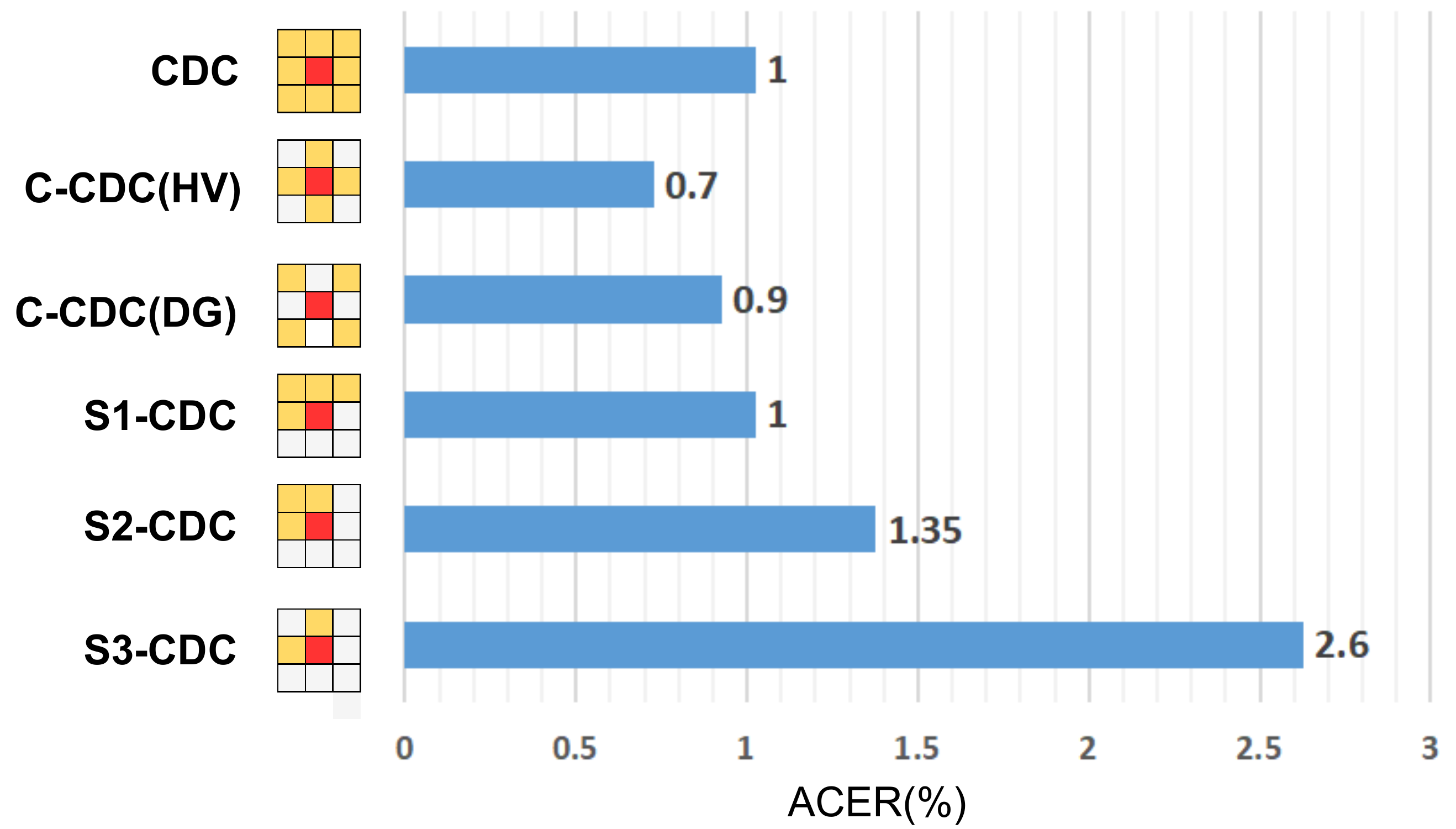}
\vspace{-1.5em}
  \caption{\small{
 Ablation study about the sparsity and neighbor locations of CDC on Protocol 1 of OULU-NPU. The neighbors with the color yellow are sampled for central difference convolution. Only the best results with optimal $\theta$ are shown.
  }
  }
 
\label{fig:CCDC}
\vspace{-0.5em}
\end{figure}

\section*{A. Impacts of Sparsity and Neighbor Locations of C-CDC}

As already shown in the Section 4.3, C-CDC(HV) and C-CDC(DG) adopt the symmetric sparse local neighbor sampling and even achieve better performance than CDC. Thus, two questions arise: 1) the sparser, the better? 2) is symmetry helpful for feature representation?  

In order to tackle these questions, we compare C-CDC with three CDC with sparse/asymmetric neighbor sampling (i.e., ``S1-CDC", ``S2-CDC", ``S3-CDC"). The results intra-tested on Protocol 1 of OULU-NPU are illustrated in Figure ~\ref{fig:CCDC}. In terms of the sparsity, the ``S2-CDC" and ``S3-CDC" respectively consider only three and two local neighbors for sampling, which is much sparser than both CDC and C-CDC. However, they suffer from sharp performance drops due to the limited local representation capacity. As for the symmetry, we compare the C-CDC with ``S1-CDC" as the latter has the same sparsity but asymmetric neighbor locations. It is interesting to see that ``S1-CDC" still has comparable performance (1\% ACER) with CDC but slightly poorer results than the symmetric C-CDC(HV) and C-CDC(DG). 

Despite with such coarse examples for comparison, it still provides evidence that the designed C-CDC family is an excellent tradeoff in terms of performance and efficiency.

\begin{figure}
\centering
\includegraphics[scale=0.7]{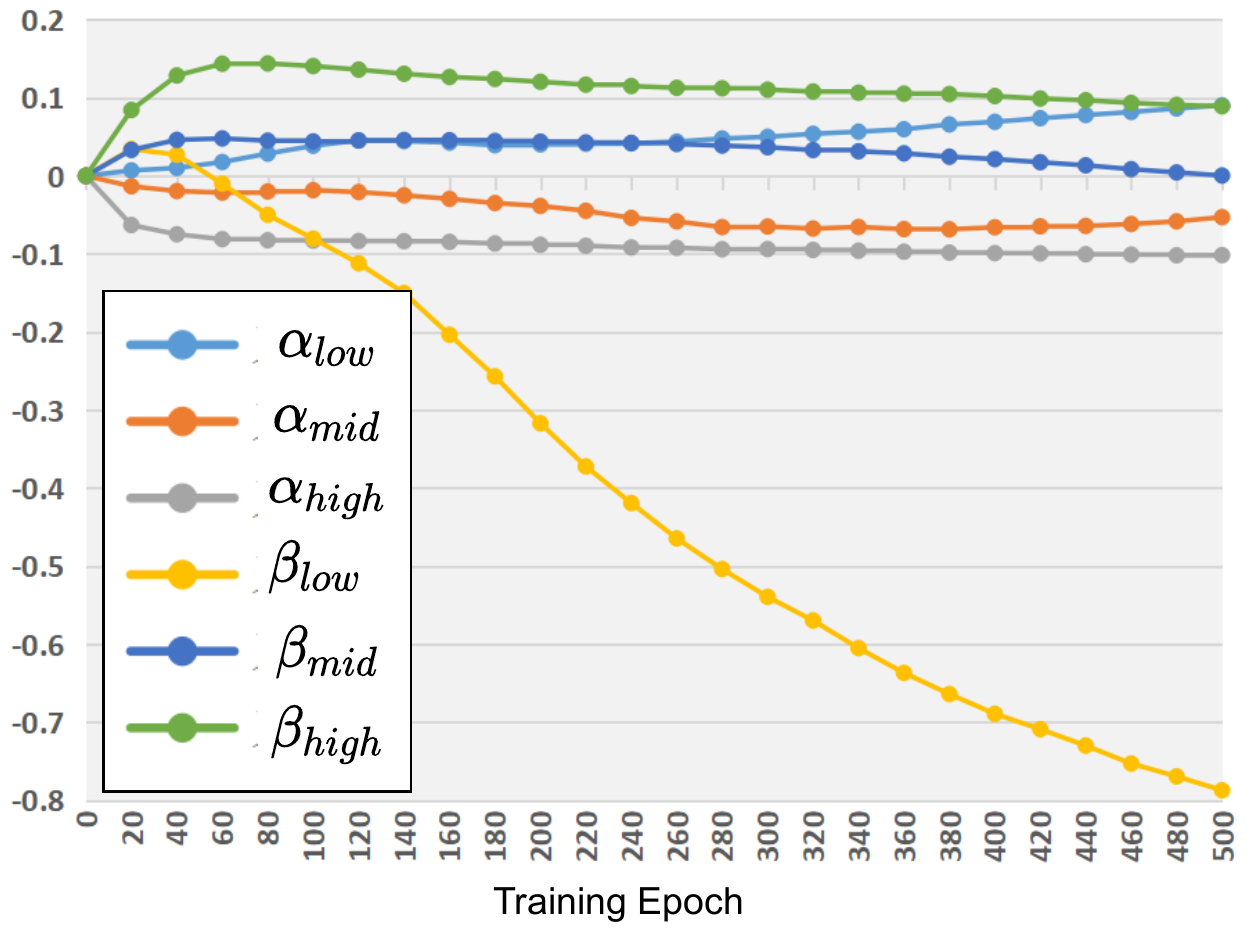}
\vspace{-1.5em}
  \caption{\small{
 Visualization of the learnable $\alpha,\beta$ of CFIM in DC-CDN during the first 500 training epochs on Protocol 1 of OULU-NPU.  
  }
  }
 
\label{fig:CFIM1}
\vspace{-0.5em}
\end{figure}

\section*{B. Discussion about CFIM}
As CFIM plays a vital role in mutual relation mining and dual-stream feature enhancement, it is necessary to explore more details about CFIM, e.g., 1) learned $\alpha,\beta$ visualization; 2) impacts of learnable/fixed setting and initialization. 

As illustrated in Figure~\ref{fig:CFIM1}, with the defaulted initialization to all zeros (i.e., $\alpha,\beta=0$, $\varphi(\alpha),\varphi(\beta)=0.5$), the trends of $\alpha_{low}$, $\alpha_{mid}$, $\alpha_{high}$, $\beta_{mid}$, $\beta_{high}$ are relatively steady except $\beta_{low}$. To be specific, $\alpha_{low}$, $\beta_{mid}$, $\beta_{high}$ prefer to hold the positive values while $\alpha_{mid}$, $\alpha_{high}$, $\beta_{low}$ keep the negative values in most times. The dramatic decrease of $\beta_{low}$ denotes that contributions from the low-level features of the C-CDC(DG) stream might be small in DC-CDN. 

\begin{figure}[t]
\centering
\includegraphics[scale=0.46]{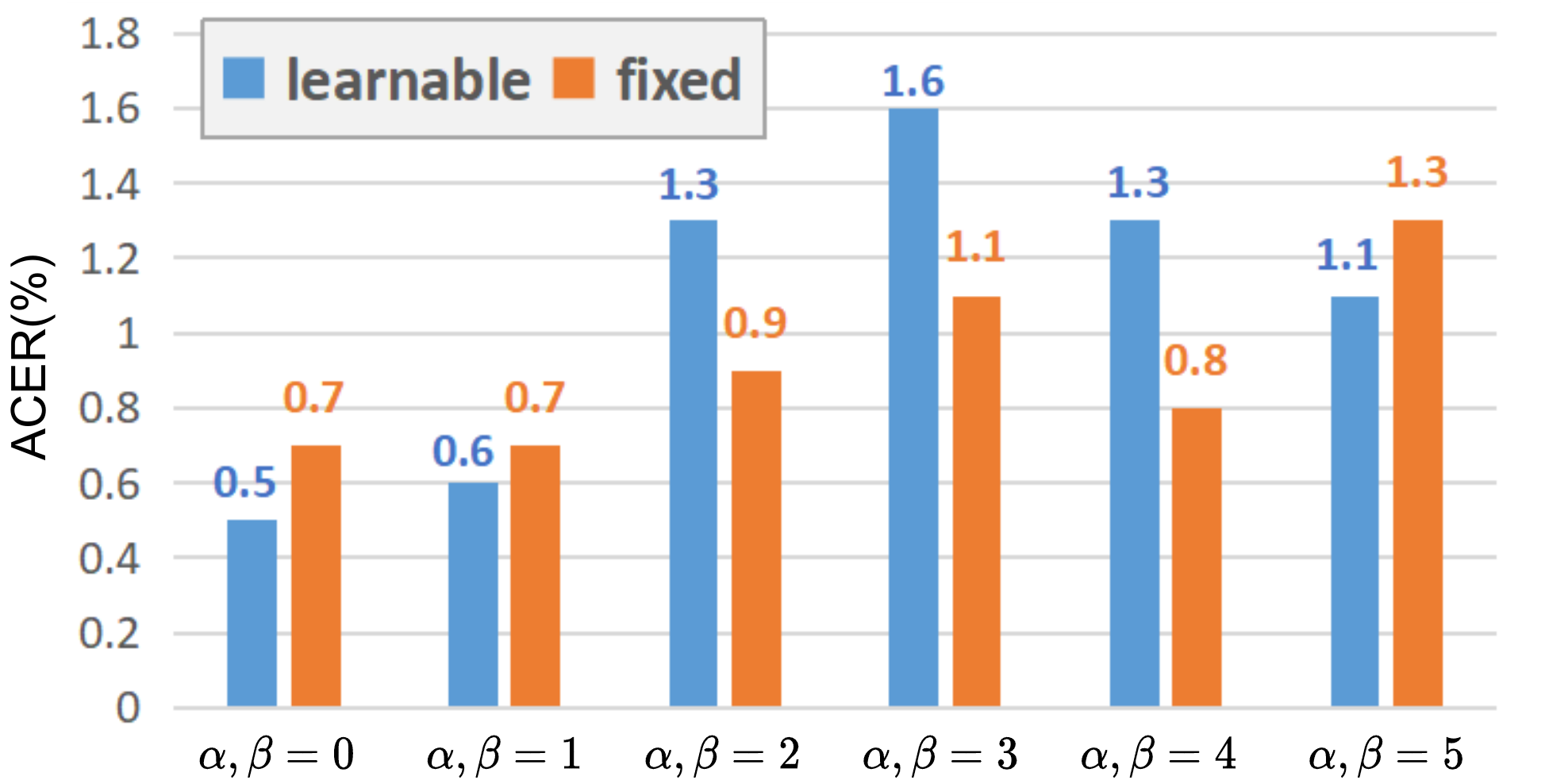}
\vspace{-1.5em}
  \caption{\small{
  Comparison of the learnable and fixed settings of $\alpha,\beta$ in CFIM with different initialized values on Protocol 1 of OULU-NPU. 
  }
  }
 
\label{fig:CFIM2}
\vspace{-0.5em}
\end{figure}

Moreover, it can be seen from Figure~\ref{fig:CFIM2} that the initialization values influence the CFIM a lot especially with learnable setting. Small initial values (e.g., $\alpha,\beta=[0,1]$) seem to be more suitable for subsequent dual-stream feature learning. The best performance could be achieved (0.5\% ACER in Protocol 1 of OULU-NPU) when $\alpha,\beta$ are learnable and initialized as zeros.

\section*{C. Analysis of PE Augmentation}
In the paper, although the PE Augmentation is proven to be helpful for DC-CDN performance improvement, it is still necessary to evaluate its generalization ability for different architectures under various scenarios. Besides DC-CDN, here we also give comparisons of two classical FAS architectures, i.e., DepthNet~\cite{Liu2018Learning}, CDCN~\cite{yu2020searching} on intra- and cross-type testings. As shown in Figure~\ref{fig:PE}, PE augmentation consistently improves DepthNet and CDCN when the training data are even small-scale (Protocol 4 of OULU-NPU) or testing under unseen attack types (SiW-M). All these results demonstrate that PE augmentation is promising for the FAS task.  

\begin{figure}[t]
\centering
\includegraphics[scale=0.32]{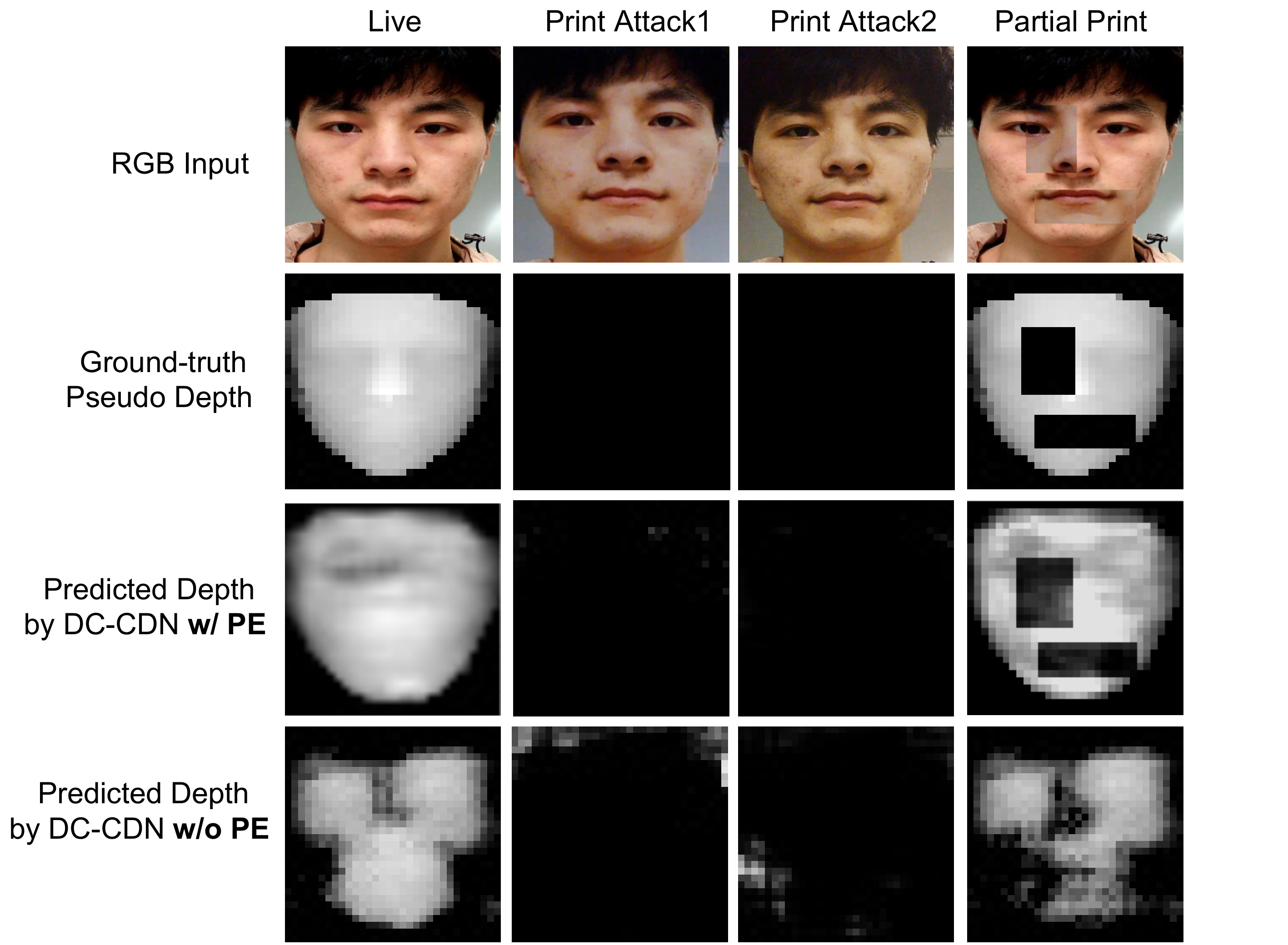}
\vspace{-1.5em}
  \caption{\small{
  Visualization of the predicted results from DC-CDN w/ and w/o PE augmentation. When given the inputs with partial print attacks, the model trained w/ PE is able to discover the spoofing patches while that trained w/o PE fails.
  }
  }
 
\label{fig:PE_v}
\vspace{-0.5em}
\end{figure}

As partial attacks (e.g., partial print, mask and funny eye glasses) have been used for hacking the face recognition system, here we show a visualization about the predictions from DC-CDN trained w/ and w/o PE augmentation. The models are trained and tested on the sub-Protocol 4-1 of OULU-NPU. As illustrated in the first three columns of Figure~\ref{fig:PE_v}, both the models (trained w/ or w/o PE augmentation) perform well for the normal live face and print attacks. However, we can see from the last column that w/o PE augmentation, it fails to capture the spoofing regions from the synthesized face with partial attacks. In contrast, the model trained w/ PE could clearly discover and localize them.

\begin{figure}[t]
\centering
\includegraphics[scale=0.5]{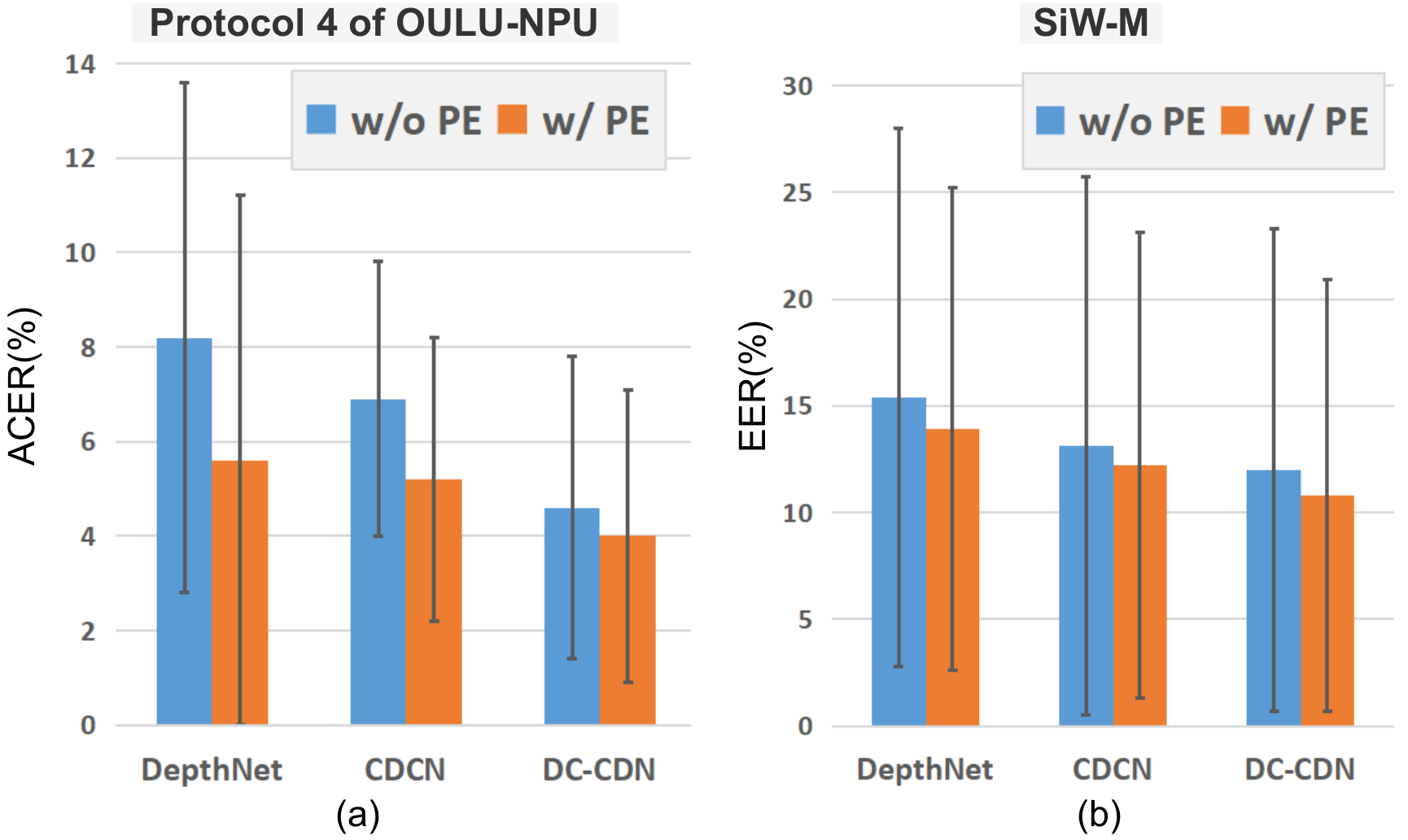}
\vspace{-1.5em}
  \caption{\small{
  Ablation study about the PE augmentation for different architectures and testing protocols.
  }
  }
 
\label{fig:PE}
\vspace{-0.5em}
\end{figure}

\bibliographystyle{named}
\bibliography{ijcai21}

\end{document}